\documentclass[10pt,twocolumn,letterpaper]{article}

\usepackage[pagenumbers]{cvpr} 
\usepackage{algorithm}
\usepackage{algpseudocode}
\usepackage{amsmath}
\usepackage{colortbl} 
\usepackage{multirow} 
\usepackage{tabularx}
\usepackage{float}

\usepackage[dvipsnames,table,xcdraw]{xcolor}
\definecolor{cvprblue}{rgb}{0.21,0.49,0.74}
\usepackage[pagebackref,breaklinks,colorlinks,citecolor=cvprblue]{hyperref}

\def\paperID{xxxx} 
\def\confName{CVPR}
\def\confYear{2025}

\title{Facial Dynamics in Video: Instruction Tuning for Improved Facial Expression Perception and Contextual Awareness}

\author{
Jiaxing Zhao$^{1}$\thanks{Equal contribution.}\quad
Boyuan Sun$^{2,1}$\samethanks \quad
Xiang Chen$^{1}$\quad
Xihan Wei$^1$\quad \\
{\normalsize{$^1$Tongyi Group, Alibaba}}\quad
{\normalsize{$^2$VCIP, CS, Nankai University}}\\
{\tt{\small{\{zjx244036,xchen.cx,xihan.wxh\}@alibaba-inc.com,}}}\\
{\tt{\small{boyuansun@mail.nankai.edu.cn,}}} \\
\textbf{\normalsize{\url{https://github.com/Jiaxing-star/FacialDynamic}}}\vspace{-3mm}}

\newcommand*\samethanks[1][\value{footnote}]{\footnotemark[#1]}

\begin{document}
\maketitle
\begin{abstract}
Facial expression captioning has found widespread application across various domains.
Recently, the emergence of video Multimodal Large Language Models (MLLMs) has shown promise in general video understanding tasks. 
However, describing facial expressions within videos poses two major challenges for these models: (1) the lack of adequate datasets and benchmarks, and (2) the limited visual token capacity of video MLLMs.
To address these issues, this paper introduces a new instruction-following dataset tailored for dynamic facial expression caption. 
The dataset comprises 5,033 high-quality video clips annotated manually, containing over 700,000 tokens. 
Its purpose is to improve the capability of video MLLMs to discern subtle facial nuances.
Furthermore, we propose FaceTrack-MM, which leverages a limited number of tokens to encode the main character's face. 
This model demonstrates superior performance in tracking faces and focusing on the facial expressions of the main characters, even in intricate multi-person scenarios.
Additionally, we introduce a novel evaluation metric combining event extraction, relation classification, and the longest common subsequence (LCS) algorithm to assess the content consistency and temporal sequence consistency of generated text.
Moreover, we present FEC-Bench, a benchmark designed to assess the performance of existing video MLLMs in this specific task.
All data and source code will be made publicly available.
\end{abstract}    
\section{Introduction}
\label{sec:intro}

Facial expressions play a crucial role in daily communication and are a vital component of human interaction. Their significance has led to growing interest from researchers in fields such as human-computer interaction (HCI)~\cite{thefeatureofemotion}, driving assistance~\cite{Driveremotion}, and mental health~\cite{healthemotion}.
As researchers have shifted their focus from static images to dynamic video content, Dynamic Facial Expression Recognition (DFER)~\cite{yu2018spatio, chen_9209166_2023, liu2020saanet, chen2024finecliper, li2024dual, foteinopoulou_emoclip_2024} has attracted attention from experts in psychology, computer science, linguistics, neuroscience, and related disciplines due to its wide-ranging applications. The goal of DFER~\cite{lucey2010extended, valstar2010induced, zhao2011facial, jiang2020dfewlargescaledatabaserecognizing} is to classify video sequences into one of several fundamental emotional categories, including neutral, happiness, sadness, surprise, fear, disgust, and anger.
However, in practical applications, characters in a video often exhibit a range of emotions at different stages, making discrete emotion classification insufficient for accurately captioning facial expressions. Even multi-class classification fails to adequately capture the subtleties of facial changes.

\begin{figure}
    \vspace{-5mm}
    \centering
    \includegraphics[width=1\linewidth]{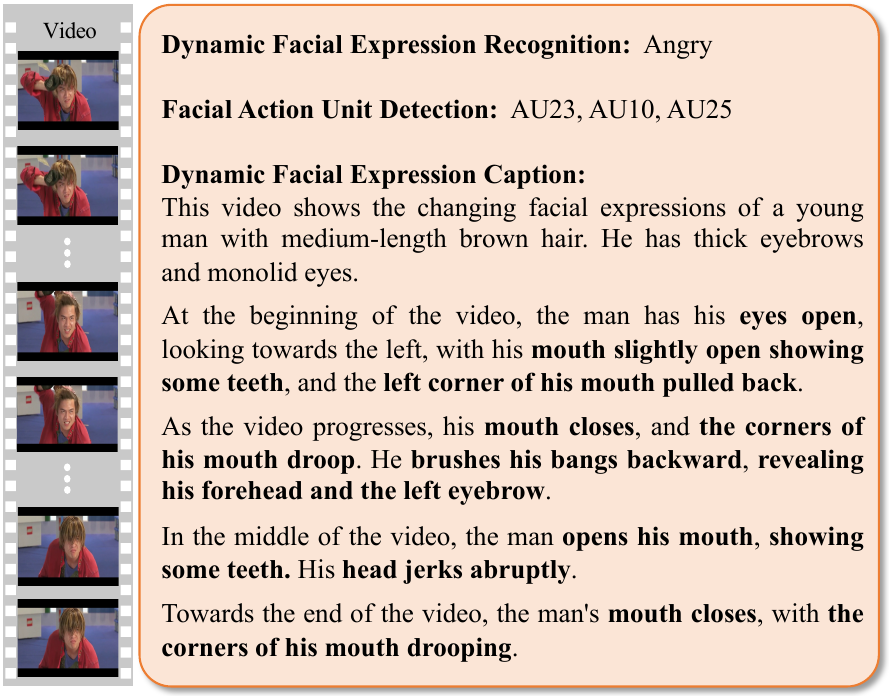}
    \vspace{-0.5cm}
    \caption{Comparison of annotation styles between different facial expression tasks. Among these, DFER only includes one selected fundamental emotional category, whereas FAUD contains several specific action units. Our DFEC includes detailed facial movements described using natural language.
    \vspace{-0.3cm}}
    \label{fig:teaser} 
    \vspace{-0.3cm}
\end{figure}
To address this limitation, we introduce a new task called Dynamic Facial Expression Caption (DFEC). The goal of DFEC is to use natural language to fluently and accurately describe the facial expressions of main characters in videos.
In this task, the model is required to generate content that covers all significant facial expressions of the main characters in the video without introducing hallucinations, especially when the video contains subtle, rapid, and meaningful changes in expressions, which increases the challenge.
Specifically, we leverage the content provided by facial units to describe the facial information of characters in videos. 
Unlike Facial Action Unit Detection (FAUD)~\cite{zhao2016deep, shao2018deep, luo2022learning, romero2018multi, ning2024representation, yuan2024auformer}, which focuses on identifying specific action units, our approach employs temporally coherent natural language, which is more aligned with downstream applications and better suited for large language models. 
As illustrated in Figure~\ref{fig:teaser}, compared to existing DFER and Facial Action Unit Detection (FAUD) tasks, our method provides more detailed and natural captions of facial expressions in videos. 
In this paper, we construct a high-quality DFEC dataset, FDA, containing 5,033 manually annotated videos to help video MLLMs better understand facial expressions in videos.

In recent years, MLLMs~\cite{chatgpt, openai2023gpt4, alayrac2022flamingo,zhu2023minigpt,ye2023mplugowl,bai2023qwen,chen2023internvl,dong2024internlm} have achieved significant success in the perception and understanding of images. Consequently, the perception and understanding of videos have garnered increasing attention from researchers ~\cite{damonlpsg2024videollama2,zhao2025llava,xu2024pllava,zhang2024llavanext-video,wang2024tarsierrecipestrainingevaluating, li2023videochat,maaz2023video,lin2023video,wang2024internvideo2,liu2024world}.
However, when dealing with video inputs, the need to process multiple frames results in a limited number of tokens being allocated to each frame. 
Even with the highly advanced Qwen2-VL~\cite{Qwen2VL} model, when processing video input, the encoding for each frame is limited to only 138 tokens. 
This token limitation often proves insufficient for encoding detailed information. In videos, the facial region usually occupies only a small portion of the frame, and this lack of detailed encoding capability can significantly impact the performance of our DFEC task.

In this paper, we introduce a novel MLLM named FaceTrack-MM, which is designed to track faces in videos and focus on the main characters' facial expressions in multi-person scenes.
Specifically, we integrate a dynamic video face tracking module into the MLLM to accurately model the facial regions of the main characters in videos. 
The extracted features are then projected and fed into the LLM to generate more precise and detailed captions of the facial region.
Besides, We propose a novel evaluation metric Temporal Event Matching (TEM), which combines event extraction, relation classification, and the Longest Common Subsequence (LCS) algorithm to assess the semantic consistency and temporal sequence of generated text.
We randomly selected 1,000 samples from the annotated data to form the test set. Using this test set, we constructed a facial expression caption benchmark, FEC-Bench, to compare the performance of 15 open-source and proprietary models across 8 metrics.

We summarize the key contributions of our paper as follows:
\begin{enumerate}
    \item We construct a high-quality DFEC dataset containing 5,033 manually annotated samples to help video MLLMs better understand and describe facial expressions in videos.
    \item We propose a novel FaceTrack-MM, which significantly improves the ability of existing large models to encode facial details.
    \item We introduce a new evaluation method, Temporal Event Matching (TEM), to explicitly assess the content consistency and order consistency of the generated text.
    \item We construct a FEC-Bench for the DFEC task, laying the foundation for future research in this field.
\end{enumerate}

\section{Related Work}

\begin{figure*}
      \setlength{\abovecaptionskip}{2pt}
     \centering
         \begin{subfigure}[b]{0.20\textwidth}
         \centering
         \includegraphics[width=\textwidth]{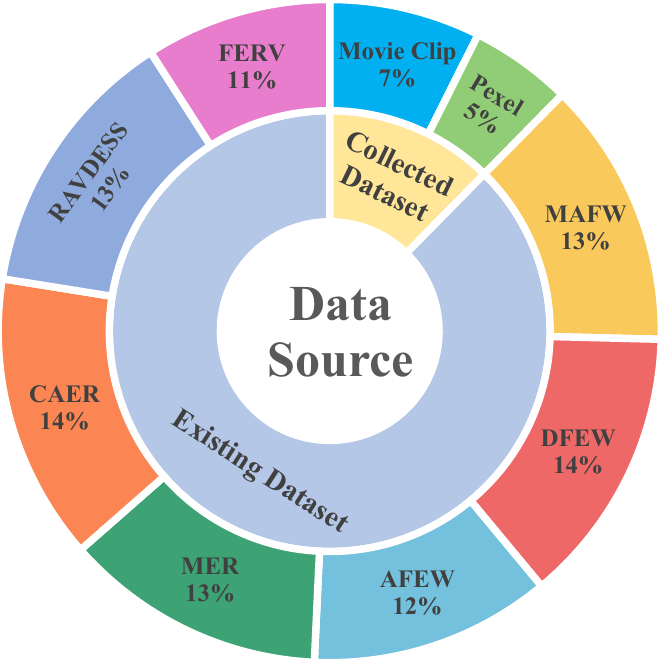}
         \caption{Data source distribution.}
         \label{fig:data source}
     \end{subfigure}
     \hfill
    \begin{subfigure}[b]{0.26\textwidth}
         \centering
         \includegraphics[width=\textwidth]{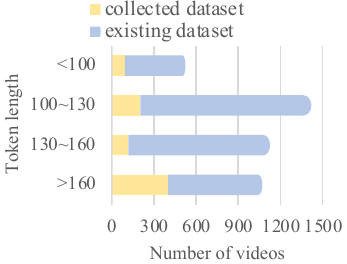}
         \caption{Token length}
         \label{fig: token length}
     \end{subfigure}
          \begin{subfigure}[b]{0.26\textwidth}
         \centering
         \includegraphics[width=\textwidth]{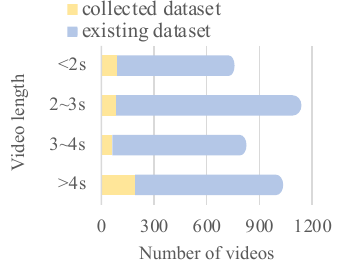}
         \caption{Video length}
         \label{fig: video length}
     \end{subfigure}
     \hfill
     \begin{subfigure}[b]{0.26\textwidth}
         \centering
         \includegraphics[width=\textwidth]{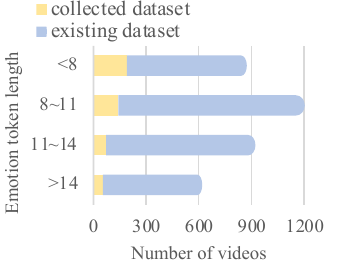}
         \caption{Emotion token length}
         \label{fig:emotion number}
     \end{subfigure}
     \hfill

        \caption{Some statistics of proposed DFEC dataset. }
        \label{fig:statistics}
    \vspace{-20pt}
\end{figure*}

\subsection{Video Caption Dataset}
Existing video captioning datasets can be categorized into three groups based on video duration. Short video datasets~\cite{chen2011collecting, rohrbach2017movie, wang2019vatex, xu2016msr} typically contain video clips ranging from 5 to 30 seconds in length. 
Longer video datasets~\cite{huang2020multimodal, krishna2017dense, zhou2018towards, mangalam2024egoschema} include videos that vary in length from 1 to 5 minutes. Very long video datasets~\cite{islam2024video, fu2024videomme, zhou2024mlvu} consist of videos that can last for hours.
However, these datasets often emphasize overall content and action recognition in videos.
Specifically, such descriptions typically include specific actions being performed (for example, ``A person is drinking coffee."), the presence of certain objects in the scene (for example, ``There is a book on the table."), or the development of a sequence of events over time (for example, ``The video begins with a man watching TV, then he stands up, walks to a table, and engages in conversation with a woman").
Despite the growing importance of facial expressions in downstream tasks such as generative models~\cite{ding2023diffusionrig, li2023photomaker} and digital human representations~\cite{vilchis2023survey, ozacar2024digihuman}, there is currently a lack of datasets that specifically focus on detailed descriptions of human facial expressions. Existing datasets often emphasize overall content and action recognition in videos, rather than the details of facial expressions.
To address this gap, we introduce a dynamic facial expression caption dataset containing 5,033 high-quality video clips, annotated manually with over 50,000 facial expression words and over 700,000 tokens. We have built a comprehensive benchmark based on this dataset, laying the foundation for further exploration in this direction.

\subsection{Multimodal Large Language Models}
Existing large models can be categorized into two primary types: proprietary large models and open large models. 
Proprietary large models are typically developed by private organizations or companies, with restricted access and usage rights, often protected by intellectual property laws. 
In contrast, open large models are designed to be accessible to the broader community, with their architectures, training data, and  the pre-trained weights, being made publicly available.

Proprietary large models, such as GPT-4V~\cite{gpt4v}, GPT-4~\cite{openai2023gpt4},  GPT-4o~\cite{openai2024gpt4o}, Gemini~\cite{geminiteam2024geminifamilyhighlycapable}, Claude-3.5~\cite{Claude2024}, Qwen2-VL~\cite{Qwen-VL,Qwen2VL}, and Qwen2.5~\cite{qwen2.5}, excel in a variety of visual scenarios. However, their capabilities are limited in specific subdomains, such as facial description. Additionally, since these models are not open-sourced, there is no way to optimize or build upon them.

In recent years, community models have also been developing rapidly, with a plethora of excellent works emerging~\cite{damonlpsg2023videollama, damonlpsg2024videollama2, li2024llava, gao2023llama, liu2023llava, liu2023llava1.5, zhang2024llavanext-video,li2024llavaone,li2024llavanext-ablations, ye2023mplug, ye2023mplugowl2,li2024llamavid, xu2024pllava}. 
LLaMA-VID~\cite{li2024llamavid} empowers existing frameworks to support hour-long videos.
LLaVA-OneVision~\cite{li2024llavaone} simultaneously push the performance boundaries of open LMMs in three important computer vision scenarios: single-image, multi-image, and video scenarios.
Video-LLaMA~\cite{damonlpsg2023videollama} proposes a multi-modal framework that empowers Large Language Models (LLMs) with the capability of understanding both visual and auditory content in the video.
Video-LLaMA2 \cite{damonlpsg2024videollama2} enhances spatial-temporal modeling and audio understanding in video and audio-oriented tasks.
mPLUG-Owl2 \cite{ye2023mplugowl2} introduces shared modules for better modality collaboration. 
BLIP-2~\cite{li2023blip2} uses Q-Former~\cite{zhang2023vision} to connect the visual and linguistic modalities. In BLIP-3~\cite{xue2024xgenmmblip3familyopen}, Q-Former is replaced by more scalable visual token samplers, such as perceptual resamplers.
However, these existing community models often focus on addressing general scenarios and also suffer from inadequate encoding of facial information, leading to suboptimal performance in tasks requiring detailed facial analysis.
In this paper, we propose a video MLLMs named FaceTrack-MM, which aims to address the shortcomings of existing models in facial encoding.

\section{Dataset}
\label{sec:dataset}
In this paper, we construct a Dynamic Facial Expression Caption dataset, FDA, to bridge the gap between the broader video understanding research community, which has traditionally focused on narrative progression and main content, and the downstream applications that greatly benefit from detailed captions of facial changes in videos.
In this section, we first introduce the sources of the data in our dataset, then describe the annotation process, followed by an analysis of the basic properties of the dataset. Finally, we propose the Temporal Event Matching (TEM), a new evaluation metric for long text captions.

\begin{figure*}
    \vspace{-5mm}
    \centering
    \includegraphics[width=1\linewidth]{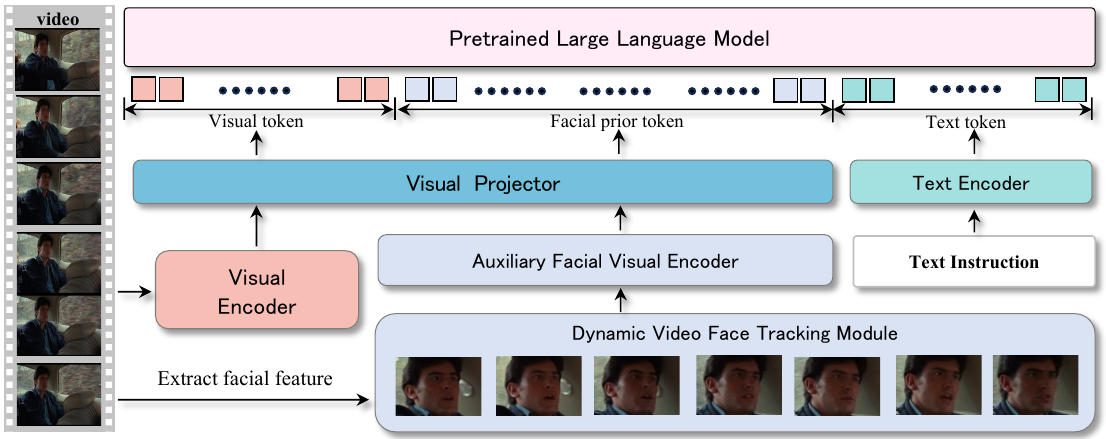}
    \vspace{-0.6cm}
    \caption{\textbf{Architecture of FaceTrack-MM.} Our FaceTrack-MM leverages FaceXFormer~\cite{narayan2024facexformer} as the auxiliary facial visual encoder to extract facial features of the main characters and uses CLIP-ViT-Large~\cite{radford2021clip} as the visual encoder. We utilize the STC module~\cite{damonlpsg2024videollama2} as the visual projector to inject temporal information and use Mistral-7B-Instruct~\cite{jiang2024mixtralexperts} for the pretrained large language model.\vspace{-0.5cm}}
    \label{fig:pipeline} 
\end{figure*}

\subsection{Data Sources}
Our data sources are primarily divided into two parts. The first part is extracted from existing datasets, specifically from video datasets related to emotions, to ensure a rich variety of facial expressions. 
We refer to this as the existing dataset. 
The second part consists of self-collected data, which we obtained through web scraping. 
We refer to this as the collected dataset. 

Specially, to ensure diversity in the dataset, we sampled 500 videos each from existing datasets such as MAFW~\cite{liu2023mafwlargescalemultimodalcompound}, DFEW~\cite{jiang2020dfewlargescaledatabaserecognizing},
AFEW~\cite{KOSSAIFI201723}, MER~\cite{GomezCanon2021SPM}, CAER~\cite{lee2019context}, FERV39K~\cite{wang2022ferv39klargescalemultiscenedataset}, and RAVDESS~\cite{Ravedess}. 
Furthermore, during the dataset sampling process, we perform uniform sampling based on emotion categories to maintain an approximately equal distribution of videos across different emotions. 
 Additionally, we collected 500 videos from Pexels Videos using keywords such as ``man", ``girl", ``family", and ``woman", and 500 videos from Cesituwang. Each video is no longer than 20 seconds. 
 We processed these sampled videos by removing the header and footer information and any irrelevant content.
Furthermore, we randomly extracted video clips from a large number of famous movies. 
For each clip, we performed face tracking to ensure that at least one face was present and that the face region occupied more than 5 percent of the frame. We discarded any clips that did not meet these criteria. Ultimately, we extracted 1,000 clips that satisfied the requirements.
Subsequently, we conducted a manual screening process to remove videos that contained too many people or static scenes with no movement.
We present the statistical data on the sources in Figure~\ref{fig:data source}, including the proportions of data extracted from each existing dataset and the proportions from each self-collected dataset.

\subsection{Annotation Process}
Our annotation process is primarily divided into two parts: the first part is the generation of preliminary annotations, and the second part is manual correction.  
We utilized carefully crafted prompts with GPT-4~\cite{openai2024gpt4o} to generate preliminary annotations. 
The generated preliminary annotations lack detailed captions, especially in facial changes, and contain a significant amount of hallucination. 
The prompts we used and the pre-annotated examples generated can be found in the supplementary material.

During the manual correction process, each video clip is reviewed by three annotators and finally consolidated by a final reviewer. The annotators refer to the pre-generated annotations and modify them to better meet our downstream requirements. Specifically, our annotation content is divided into three parts.

\textbf{External Attributes.} We described the external attributes of the individuals in the video, including subtle facial features.

\textbf{Facial Changes.} We focused on detailing the facial changes of the main characters in the video. In this section, we retained certain reasonable inferences generated during the pre-annotation phase, as we believe these inferences enhance the richness of the descriptions. Since these inferences are not objectively present in the video, we marked them with special identifiers.

\textbf{Summary Sentence.} Following the annotation style of the MAFW~\cite{liu2023mafwlargescalemultimodalcompound} dataset, we used a single sentence to summarize the content of the video. 

During the annotation process, annotators filtered out videos based on the following criteria: videos with excessive facial changes that made annotation difficult, videos with minimal facial changes, videos containing violent content, and videos with frequent perspective switches. Ultimately, the total number of videos remaining was 5,033.
Overall, our dataset explicitly distinguishes between objective facial descriptions and slightly subjective speculative descriptions, while also including a concise one-sentence summary that captures the content. This allows us to select the appropriate annotated content for fine-tuning the model based on our application needs. In the subsequent instruction tuning, dataset attributes, experiments, and benchmark, we use data that includes individual attributes and annotations of the main character's facial changes, with subjective inferences removed.

\subsection{Dataset Attributes}
As mentioned earlier, our data can be categorized into two types based on the source: existing data and self-collected data. Here, we analyze the properties of these two types of data. Comparative analyses with other datasets are provided in the supplementary materials. In Figure~\ref{fig: token length}, we present the statistics of the number of annotated tokens in the dataset. It can be observed that the average number of tokens in the existing data is 134.7, while the average number of tokens in the self-collected data is 125.3.

In Figure~\ref{fig: video length}, we present the statistics of the average duration of videos from the two sources. The average duration of videos from the existing data is 3.13 seconds, while the average duration of videos from the self-collected data is 4.22 seconds. This video length is efficient for capturing subtle facial changes, reducing the computational burden of data processing, facilitating annotation and model training, and making it suitable for real-time applications and emotion recognition. 

We used ChatGPT~\cite{chatgpt} to extract facial action-related descriptors from the annotations, such as ``tilted head", ``eyes widened", ``smile", etc. Then, we counted the number of keywords in each video to illustrate the level of detail in our annotations regarding facial actions. As shown in Figure~\ref{fig:emotion number}, the average number of facial action keywords in the existing data is 10.9, while in the self-collected data, it is 9.1. This demonstrates that our annotations for facial actions are highly detailed. 

\subsection{Temporal Event Matching}
Traditional evaluation metrics based on n-grams, such as CIDEr~\cite{vedantam2015cider}, struggle when assessing long video descriptions. This is because there are numerous ways to convey the same meaning in extended text, and many of these equivalent descriptions may have minimal n-gram overlap with the reference text. Moreover, manually assigning quality scores to descriptions is both labor-intensive and subjective. To address these issues, Maaz et al.~\cite{maaz2024videochatgptdetailedvideounderstanding} propose using ChatGPT~\cite{chatgpt} to rate descriptions on a scale from 1 to 5. However, the specific meanings of these ratings are ambiguous, and the ratings themselves have not been standardized.

Recently, Tarsier \cite{wang2024tarsierrecipestrainingevaluating} introduced an evaluation method called AutoDQ, which extracts text descriptions into a set of events and then calculates precision and recall by evaluating the relationships between the generated text and the reference event set, and vice versa, ultimately deriving the F-measure. 
While AutoDQ makes significant progress in capturing the semantic consistency and completeness of events, it does not account for the order of events. In many applications, especially those involving temporal and causal relationships, the order of events is crucial. Incorrect event ordering can lead to logically incoherent text, affecting the overall quality. To overcome these limitations, we propose a new evaluation metric named Temporal Event Matching (TEM) that combines event extraction, relation classification, and the longest common subsequence (LCS) algorithm to assess the semantic consistency and event ordering of generated text. Our method calculates the LCS score, normalizes it to a range of 0 to 1, and then averages it with the F-measure~\cite{wang2024tarsierrecipestrainingevaluating} of generated events and reference events to provide a comprehensive evaluation result. By integrating these components, TEM ensures that the generated text not only captures the correct events and their relationships but also maintains the correct order and coherence, providing a more holistic assessment standard for long video captions.

\begin{algorithm}
\caption{Temporal Event Matching} 
\label{alg:evaluation_metric}  
\begin{algorithmic}[1] 
\Require Generated Text $G$, Reference Text $R$ 
\Ensure Evaluation Score $S$

\State \textbf{Event Extraction:} Extract events from $G$ and $R$ to get $E_G$ and $E_R$

\State \textbf{Relation Classification:} Classify relations in $E_G$ and $E_R$ 

\Function{LCS-Score}{$E_G, E_R$}
    
    \State \textbf{Initialization:} Initialize $m \gets |E_G|$, $n \gets |E_R|$
    
    \State \textbf{Array Creation:} Create a 2D array $L$ of size $(m+1) \times (n+1)$

    \For{$i \gets 0$ to $m$}
        \State \textbf{Column Initialization:} $L[i][0] \gets 0$
    \EndFor

    \For{$j \gets 0$ to $n$}
        \State \textbf{Row Initialization:} $L[0][j] \gets 0$
    \EndFor

    \For{$i \gets 1$ to $m$}
        \For{$j \gets 1$ to $n$}
            \If{$E_G[i-1] == E_R[j-1]$}
                \State \textbf{Match Found:} $L[i][j] \gets L[i-1][j-1] + 1$
            \Else
                \State \textbf{No Match:} $L[i][j] \gets \max(L[i-1][j], L[i][j-1])$
            \EndIf
        \EndFor
    \EndFor

    \State \textbf{Return Normalized LCS Score:} \Return $\frac{L[m][n]}{m}$
\EndFunction

\State \textbf{Calculate LCS Score:} $lcs \gets \Call{LCS-Score}{E_G, E_R}$

\State \textbf{F-Measure Calculation:} Calculate precision, recall, and F-measure using $E_G$ and $E_R$

\State \textbf{Comprehensive Evaluation Score:} $S \gets \frac{lcs + F\text{-measure}}{2}$

\State \textbf{Return Final Score:} \Return $S$
\end{algorithmic}
\end{algorithm}

As shown in Algorithm~\ref{alg:evaluation_metric}, our evaluation metric calculation is structured into five key steps. In the first step, drawing on the methodology of AutoDQ, we utilize the ChatGPT model to extract a set of events, denoted as $E_G$ from the generated text and $E_R$ from the reference text, respectively. This initial extraction process aims to identify and isolate the critical events within both texts, setting the stage for subsequent comparison and assessment.

In the second step, we leverage ChatGPT once more to evaluate the relationship between each event in $E_G$ and each event in $E_R$. These relationships are categorized into three primary types: Same Meaning, Opposite Meaning, and No Relation. By classifying these relationships, we can achieve a more precise understanding of the semantic alignment between the generated and reference texts.

For the third step, to quantify the similarity between the two sets of events, we employ the Longest Common Subsequence (LCS) algorithm, a dynamic programming technique that efficiently determines the proportion of the longest subsequence of $E_R$ found within $E_G$. This ratio, referred to as the LCS score, serves as an indicator of the consistency between the generated and reference texts at the event level.

In the fourth step, we calculate the F-measure~\cite{wang2024tarsierrecipestrainingevaluating}, a composite metric that balances precision and recall, offering a comprehensive view of performance. Finally, we integrate the LCS score with the F-measure to derive a combined score, which stands as our ultimate evaluation criterion for the quality of the generated text. The specific prompts used throughout this process are detailed in the supplementary materials.

\section{Method}
\definecolor{lightlightgray}{gray}{0.95}

In this section, we first introduce the motivation behind our work. We then describe the architecture of our model and finally detail the training process and implementation specifics.

\subsection{Motivation}
Since existing video large language models primarily focus on understanding the main content and narrative progression of videos, their ability to encode detailed video information is limited. Directly fine-tuning these models is insufficient for adequately modeling facial regions in videos. 

In \cite{li2024facialaffectivebehavioranalysis}, Li \textit{et al.} designed a facial prior expert to provide facial feature encoding for large language models.
Inspired by this work, we propose a method that accurately models facial regions in videos, resulting in precise facial description results.

\subsection{Architecture}
\subsubsection{Overall}
As shown in Figure~\ref{fig:teaser}, our model primarily consists of a pre-trained visual encoder, a visual projector, a dynamic face extraction module, an auxiliary facial visual encoder, and a large language decoder. 
Specifically, during the training process, for the input video, we first use a pre-trained image encoder to extract features from the selected frames. Then, we use the STC \cite{damonlpsg2024videollama2} module to map these features to the text domain. 
Additionally, for the input video, we use the dynamic face extraction module to extract facial information of the main characters. These facial features are subsequently processed by an auxiliary facial visual encoder to extract visual features. 
These visual features are also mapped to the text domain.
Finally, the visual features, prior features, and text features are combined and fed into the large-scale language model for text generation.

The pre-trained visual encoder, visual projector, and large-scale language model all follow the design of VideoLLaMA2 \cite{damonlpsg2024videollama2}. The visual encoder uses CLIP-ViT-Large \cite{radford2021clip}, the visual projector is the STC module, and the large-scale language model employs Mistral-7B-Instruct \cite{jiang2024mixtralexperts}.
We use the FaceXFormer \cite{narayan2024facexformer} architecture, designed for face analysis, as the auxiliary facial visual encoder. 

\begin{table*}[htbp]
    \centering
    \small
    \setlength{\tabcolsep}{6.8pt} 
    \renewcommand{\arraystretch}{1.2} 
    \begin{tabular}{lcccccc|cc|cc}
        \toprule
        \multirow{2}{*}{\textbf{Method}} & \textbf{LLM}& \multicolumn{4}{c}{\textbf{VideoChatGPT Scores}} & \multicolumn{2}{c}{\textbf{N-gram Based}} &\multicolumn{2}{c}{\textbf{Event Based}} \\
        \cmidrule(lr){3-6} \cmidrule(lr){7-8}\cmidrule(lr){9-10}
        & \textbf{Size} &\textbf{Correctness} & \textbf{Detail} & \textbf{Context} & \textbf{Temporal} & \textbf{CIDEr} & \textbf{Rouge-L} & \textbf{AutoDQ} & \textbf{TEM} \\
        \midrule
        GPT4-O \cite{openai2024gpt4o} & - & 4.18 & 3.98 & 4.50 & 3.92 & 0.225 & 0.179 & 0.410 & \cellcolor{lightlightgray}\textbf{0.291} \\
        GPT4-O* \cite{openai2024gpt4o} & - & 4.22 & 3.97 & 4.48 & 3.90 & 0.264 & 0.213 & 0.432 & \cellcolor{lightlightgray}\textbf{0.303} \\
        Claude3.5-Sonnet \cite{Claude2024} & - & 4.11 & 3.97 & 4.41 & 3.85 & 0.212 & 0.197 & 0.420 & \cellcolor{lightlightgray}\textbf{0.298} \\
        Claude3.5-Sonnet* \cite{Claude2024} & - & 4.13 & 4.01 & 4.49 & 4.05 & 0.243 & 0.228 & 0.442 & \cellcolor{lightlightgray}\textbf{0.307} \\
        \cmidrule(lr){1-10} 
        VideoLLaMA \cite{damonlpsg2023videollama} & 7B & 3.60 & 3.67 & 3.84 & 3.50 & 0.189 & 0.196 & 0.303 & \cellcolor{lightlightgray}\textbf{0.199} \\
        VideoChat \cite{li2023videochat} & 7B & 3.47 & 3.52 & 3.92 & 3.38 & 0.251 & 0.192 & 0.344 & \cellcolor{lightlightgray}\textbf{0.229} \\
        VideoChat2 \cite{li2023videochat} & 7B & 3.70 & 3.56 & 4.16 & 3.52 & 0.202 & 0.229 & 0.311 & \cellcolor{lightlightgray}\textbf{0.231} \\
        Chat-UniVI \cite{jin2023chatunivi} & 7B & 3.64 & 3.63 & 4.21 & 3.61 & 0.189 & 0.231 & 0.396 & \cellcolor{lightlightgray}\textbf{0.261} \\
        LLaVA-Next-Video \cite{zhang2024llavanext-video} & 7B & 4.19 & 4.07 & 4.39 &  4.04 & 0.250 & 0.249 & 0.395 & \cellcolor{lightlightgray}\textbf{0.276} \\
        ShareGPT4Video \cite{chen2024sharegpt4video} & 7B & 4.24 & 4.13 & 4.35 &  4.09 & 0.192 & 0.205 & 0.394 & \cellcolor{lightlightgray}\textbf{0.278} \\
        LLaMA-VID \cite{li2024llamavid} & 7B & 3.95 & 4.01 & 4.22 & 3.71 & 0.195 & 0.231 & 0.339 & \cellcolor{lightlightgray}\textbf{0.241} \\
        VideoLLaMA2 \cite{damonlpsg2024videollama2} & 7B & 4.17 & 4.02 & 4.47 & 3.93 & 0.253 & 0.266 & 0.344 & \cellcolor{lightlightgray}\textbf{0.258} \\
        PLLaVA \cite{xu2024pllava} & 7B & 4.21 & 4.15 & 4.37 & 4.08 & 0.268 & 0.250 & 0.393 & \cellcolor{lightlightgray}\textbf{0.257} \\
        ST-LLM \cite{liu2023one1, liu2023one2} & 7B & 4.00 & 3.98 & 4.31 & 3.94 & 0.213 & 0.238 & 0.321 & \cellcolor{lightlightgray}\textbf{0.240} \\
        Qwen2-VL \cite{Qwen2VL} & 7B  & 4.23 & 4.16 & 4.52 & 4.02 & 0.204 & 0.233 & 0.422 & \cellcolor{lightlightgray}\textbf{0.309} \\
        Tarsier \cite{wang2024tarsierrecipestrainingevaluating} & 7B  & 3.59 & 3.50 & 4.07 & 3.41 & 0.143 & 0.185 & 0.415 & \cellcolor{lightlightgray}\textbf{0.292} \\
        LLaVA-OneVision \cite{li2024llavaone} & 7B & 3.68 & 3.47 & 4.10 & 3.42 & 0.115 & 0.165 & 0.379 & \cellcolor{lightlightgray}\textbf{0.275} \\
        
        \cmidrule(lr){1-10} 
        \textbf{Ours} & \textbf{7B}  & \textbf{4.42} & \textbf{4.30} & \textbf{4.60} & \textbf{4.26} & \textbf{0.418} & \textbf{0.473} & \textbf{0.483} & \cellcolor{lightlightgray}\textbf{0.364} \\
        \bottomrule
    \end{tabular}
    \caption{Comparison of different methods on average ChatGPT scores (Correctness, Detail, Context, Temporal, Consistency) and n-gram based metrics (CIDEr, ROUGE-L), event-based metrics (AutoDQ, our proposed TEM) for 1000 videos, categorized by model type (proprietary vs. open-source). $^*$ denotes the use of in-context learning during evaluation.}
    \vspace{-0.5cm}
    \label{tab:benchmark_results}
\end{table*}

\subsubsection{Dynamic Video Face Tracking Module}
Compared to directly using a face feature extractor to extract features from each frame, which may result in multiple faces being detected per frame and faces between frames not matching one-to-one, our dynamic video face tracking module incorporates Face Detection and Multi-Object Tracking techniques. This ensures that the facial features of the main characters are consistently and accurately tracked across the video.
The module can be broken down into the following steps:

\textbf{Video Frame Downsampling and Face Feature Extraction}. Downsample the video to a uniform frame rate of 16 fps. This step helps reduce computational load while retaining sufficient spatiotemporal information. 
Subsequently, perform face keypoint detection DaMOFD \cite{liu2023damofd} and face feature extraction TransFace \cite{dan2023transfacecalibratingtransformertraining} on each downsampled video frame.

\textbf{Multi-Object Tracking}. We applied the StrongSORT \cite{du2023strongsort} multi-object tracking algorithm to extract target trajectories. This algorithm provides robust trajectory information, including the position, size, and motion path of each target

\textbf{Face Trajectory Feature Extraction}. For each generated trajectory, we first calculate the total area occupied by the trajectory in the video. The total area is related to the temporal proportion of the face trajectory in the video and the spatial proportion in each frame. We then compare the distance matrix of the face feature vectors within the trajectory to obtain the average cosine similarity within the trajectory.

\textbf{Main Trajectory Selection}. For each trajectory, we use the total area and the average cosine similarity as its features and perform K-means clustering with 2 cluster centers, representing the main character cluster and the background character cluster, respectively.

Through this process, we can obtain one or multiple facial trajectories. 
To ensure that the extracted video frames contain the faces of the main characters while maintaining as even a distribution as possible, during the frame extraction process, if a frame does not contain the face of a main character, we replace it with the nearest frame that does contain the face of a main character.

\subsection{Instruction Tuning}
Since our data is entirely manually annotated, it is of high quality but relatively limited in quantity. We fine-tuned the model using LoRA~\cite{hu2022lora} on the VideoLLaMA2-7B \cite{damonlpsg2024videollama2} base model. We generated 100 similar instructions using ChatGPT and manually selected 20 of them, some of which are shown in the supplementary material. For each annotated video, we randomly selected one of the 20 instructions to form the instruction data.

\section{Experiments}

\begin{figure*}
    \vspace{-5mm}
    \centering
    \includegraphics[width=1\linewidth]{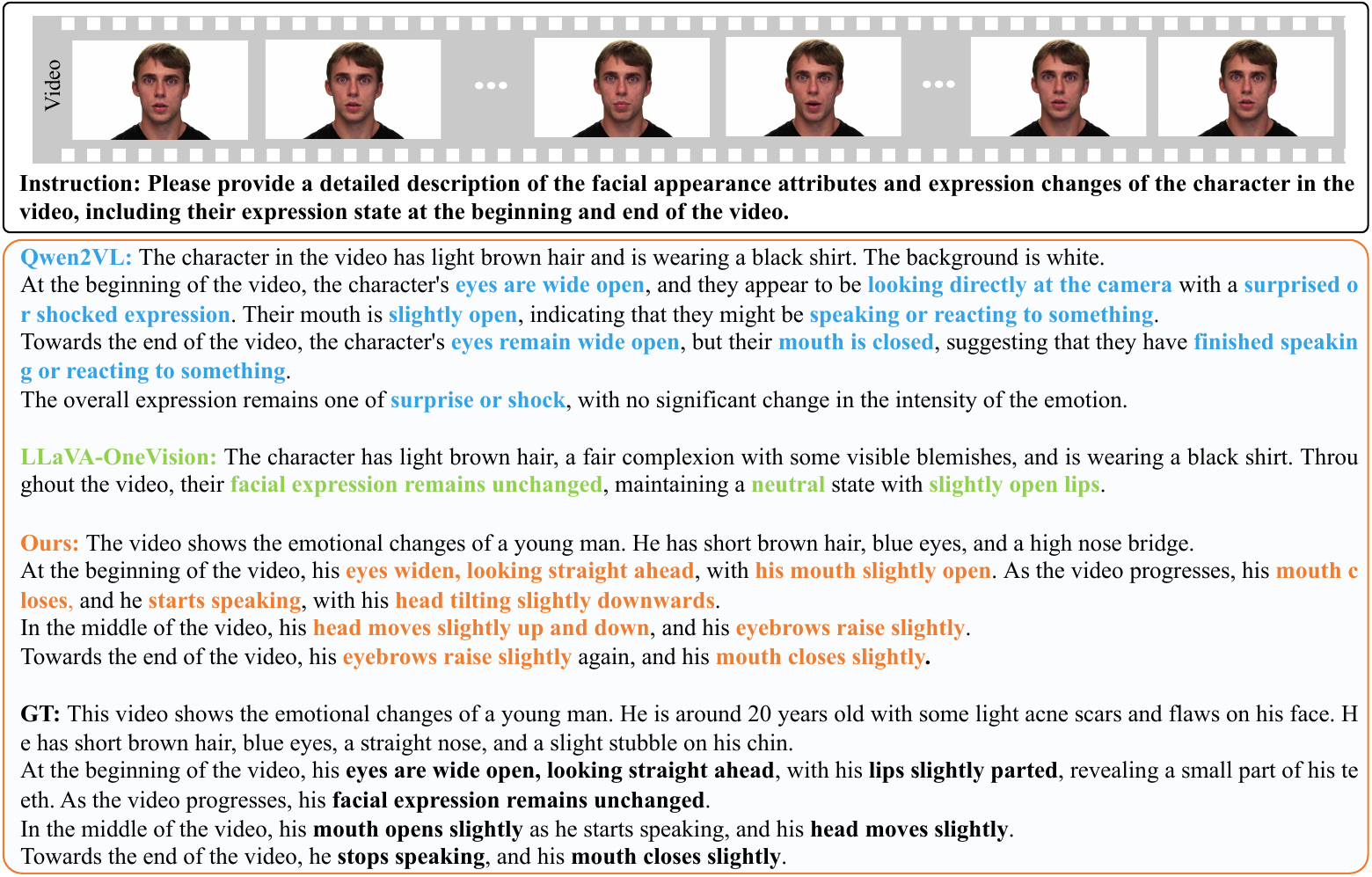}
    \vspace{-0.6cm}
    \caption{\textbf{Qualitative result comparison.} We highlight the content related to emotional and expressive changes in different methods as well as in the ground truth (GT). Our model demonstrates superior capability in capturing changes in facial expressions.\vspace{-0.3cm}}
    \label{fig:method compare} 
\end{figure*}
\subsection{Implementation Details}
Our code is based on VideoLLaMA2~\cite{damonlpsg2024videollama2}. The LoRA~\cite{hu2022lora} parameters we used are set as follows: lora\_r is set to 64, lora\_alpha is set to 128. The batch size is set to 32, and we train for 3 epochs. Each video is sampled to 16 frames. During training, we use bfloat16 precision to improve computational efficiency and model performance. The entire training process is conducted on 8 NVIDIA A100 GPUs.

\subsection{FEC-Bench}
We randomly selected 1,000 samples from our constructed dataset to form the test set. In addition to the metrics we proposed, we employed a variety of evaluation metrics, including four ChatGPT-related metrics~\cite{maaz2024videochatgptdetailedvideounderstanding}, the n-gram based metric CIDEr~\cite{vedantam2015cider}, and AutoDQ~\cite{wang2024tarsierrecipestrainingevaluating}, to comprehensively assess the performance of a large number of video MLLMs, including proprietary models.
Due to the lack of instruction tuning specifically on our dataset, the generated text does not always match the format of the reference text. This discrepancy leads to biased results when using ChatGPT for evaluation. Therefore, before evaluating the generated text using ChatGPT, we first use ChatGPT to convert the generated text into a similar format as the reference text without altering its content. The specific prompts used and the effects before and after conversion are detailed in the supplementary materials.
This extensive benchmarking not only provides a robust comparison framework but also aids in identifying the strengths and weaknesses of each model. The detailed results of this comprehensive evaluation are presented in Table \ref{tab:benchmark_results}.
We observe that even with carefully tuned prompts and context learning, existing proprietary large-scale models and open-source community large-scale models still struggle to achieve excellent performance in the task of video facial expression description. In contrast, our method, after additional modeling of the main character's facial information and instruction tuning on DFEC, significantly outperforms all other methods on the FEC-Bench.

\subsection{Ablation Study} 

In Table~\ref{tab:face_tracking_comparison}, we present the methods and their effects by gradually increasing the utilization of facial information. Our baseline is VideoLLaMA2. First, we performed instruction tuning directly on our training set using VideoLLaMA2, observing a significant improvement with around a 7 percent increase in two event-based evaluation metrics. Next, we used RetinaFace to extract and concatenate facial regions, randomly selecting 2 faces if more were detected per frame. These concatenated images were fed into a CLIP model for feature extraction, but this approach showed almost no improvement. Then, we replaced RetinaFace with our dynamic face tracking module, which significantly improved all metrics. Finally, we used a specialized encoder to extract facial features, further enhancing the results.

\begin{table}[t]
    \centering
    \renewcommand{\arraystretch}{1.1} 
    \setlength{\tabcolsep}{5pt} 
    \begin{tabular}{l|ccc}
        \toprule
        \textbf{Method} & \textbf{VC Avg.} & \textbf{AutoDQ} & \textbf{TEM} \\ 
        \cmidrule(lr){1-4}
        VideoLLaMA2~\cite{damonlpsg2024videollama2} & 4.15 & 0.344  & 0.258 \\ 
        Instruction Tuning & 4.28 & 0.422 & 0.321 \\ 
        RetinaFace Face Det & 4.31 & 0.426 & 0.324 \\ 
        Dynamic Face Tracking & 4.37 & 0.471 & 0.357 \\ 
        \textbf{FaceXFormer (ours)} & \textbf{4.40} &\textbf{ 0.483} & \textbf{0.364} \\ 
        \bottomrule
    \end{tabular}

    \caption{Ablation Study on Dynamic Video Face Tracking Module. VC Avg. represents the average VideoChatGPT score.}
    \vspace{-0.6cm}
    \label{tab:face_tracking_comparison}
\end{table}

\subsection{Qualitative Analysis}
Here, we present the visualization results of some open-source community models and our proposed model. 
As shown in Fig.~\ref{fig:method compare}, it is evident that due to data and model limitations, existing models lack the capability to provide detailed, accurate, and hallucination-free descriptions of facial expressions in videos. 
Specifically, Qwen2VL~\cite{Qwen2VL} not only misjudges emotions but also exhibits certain hallucinations, while LLaVA-OneVision~\cite{li2024llavaone} can only provide simple descriptions and overlooks changes in expressions.
In contrast, our method can accurately describe the facial expression states and changes of the main characters in the video. 
It identifies the dynamics of emotions and facial expressions, showcasing its proficiency in retrieving and interpreting visual information related to the character. These capabilities highlight our proposed method's potential to overcome the challenges associated with processing and understanding changes in human expressions.

\section{Conclusion}

In this paper, we tackle the challenges of dynamic facial expression captioning using video Multimodal Large Language Models (MLLMs). We introduce a new instruction-following dataset containing 5,033 high-quality video clips with over 700,000 tokens, designed to enhance MLLMs' ability to capture subtle facial differences. We also propose FaceTrack-MM, a model that efficiently tracks and focuses on the main character's facial expressions in complex scenes using a limited number of tokens. Additionally, we develop a novel evaluation metric that combines event extraction, relation classification, and the Longest Common Subsequence (LCS) algorithm to assess the quality of generated captions. Finally, we present FEC-Bench, a benchmark for evaluating MLLMs performance in facial expression captioning task. 
Our results demonstrate significant improvements in capturing and describing dynamic facial expressions, contributing to the advancement of this field and providing valuable resources for future research.
\clearpage

\setcounter{page}{1}
\maketitlesupplementary

\section{Generation of Preliminary Annotations}
\begin{figure*}[t]
    \centering
    \includegraphics[width=0.98\linewidth]{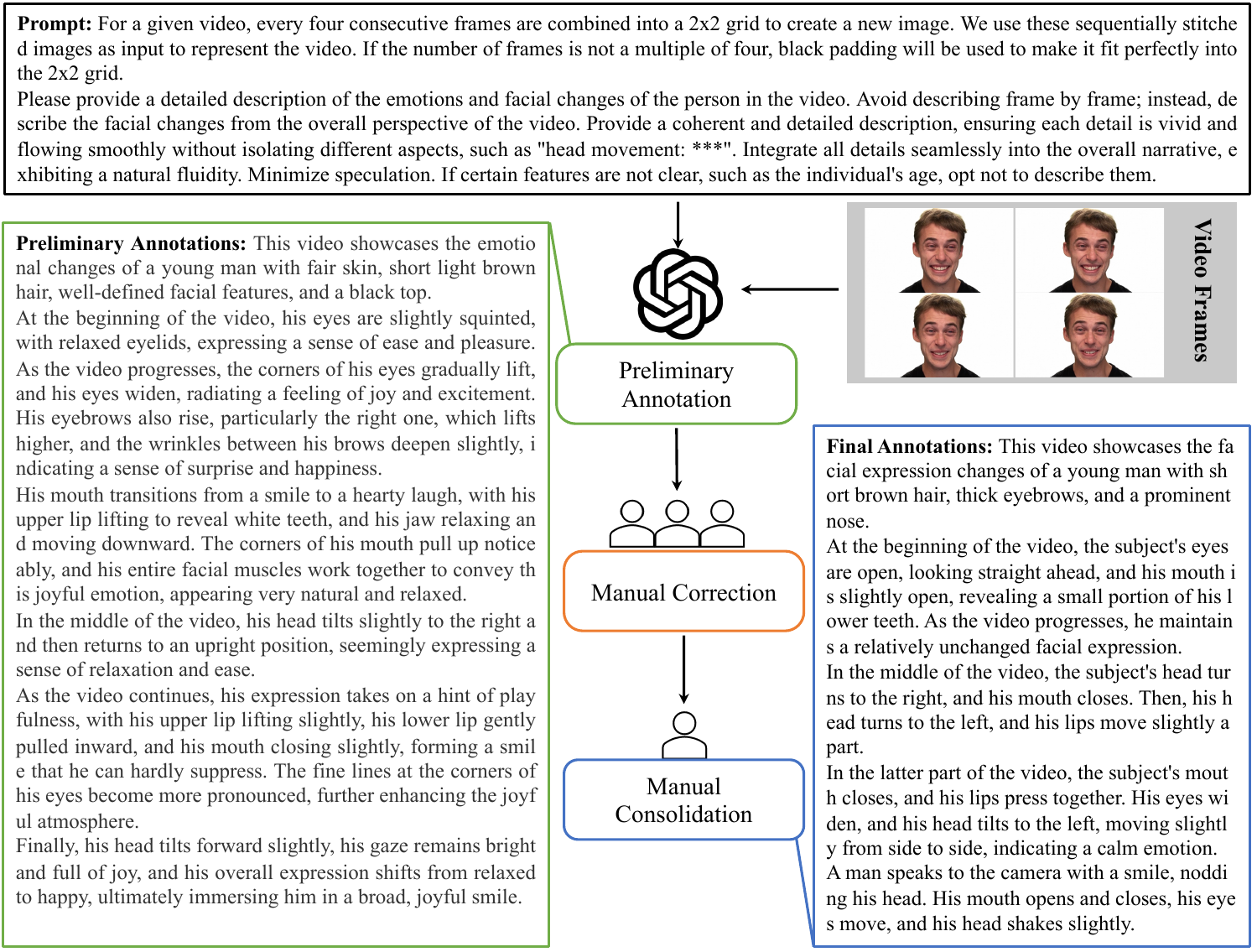}
    \vspace{-0.1cm}
    \caption{\textbf{Pipeline of the FDA annotation process. }We use ChatGPT to preliminarily annotate the emotions and facial changes of the person in the video and introduce manual correcton and consolidation to get the refined final annotations.\vspace{-0.3cm}}
    \label{fig: anno} 
\end{figure*}

In Fig.~\ref{fig: anno}, we present our comprehensive annotation process and the prompts used during the pre-generation phase. To ensure the quality and consistency of our annotations, we adopted a multi-step approach. Initially, we utilized GPT4-o~\cite{openai2024gpt4o} with carefully designed prompts to generate preliminary annotations. These prompts were crafted to guide the model in producing detailed and accurate initial annotations, taking into account the specific requirements of our dataset. After generating the preliminary annotations, we conducted a thorough manual correctness process. This involved manual modifications by three annotators who ensured that the annotations met the desired standards. The annotators also resolved any inconsistencies or ambiguities that were identified during the review. Finally, the refined annotations were reviewed again by a final reviewer to ensure the highest possible quality. The steps taken to refine the annotations is detailed in the paper.

\section{Comparison with Existing Video Expression Datasets}
In Table~\ref{tabel-comp}, we compare our dataset, FDA, with existing video expression datasets. As can be seen, the most distinctive feature of our dataset is that it does not provide predefined emotion categories for each video clip. Instead, it offers rich and detailed natural language descriptions, with an average of 10.3 expression-related annotation words per clip. This approach allows for a more nuanced and flexible representation of facial expression, capturing the subtleties and complexities that predefined categories might overlook. Additionally, the sources of our videos are more diverse, encompassing a wide range of scenarios and contexts. This diversity helps to improve the representativeness and generalization capability of the dataset.

\section{Prompt in FEC-Bench}
\subsection{Temporal Event Matching}
In constructing the evaluation metric TEM, we utilized ChatGPT~\cite{chatgpt} for relation extraction and relation classification. We provide a detailed demonstration of this process.

\begin{figure*}
    \centering
    \includegraphics[width=0.98\linewidth]{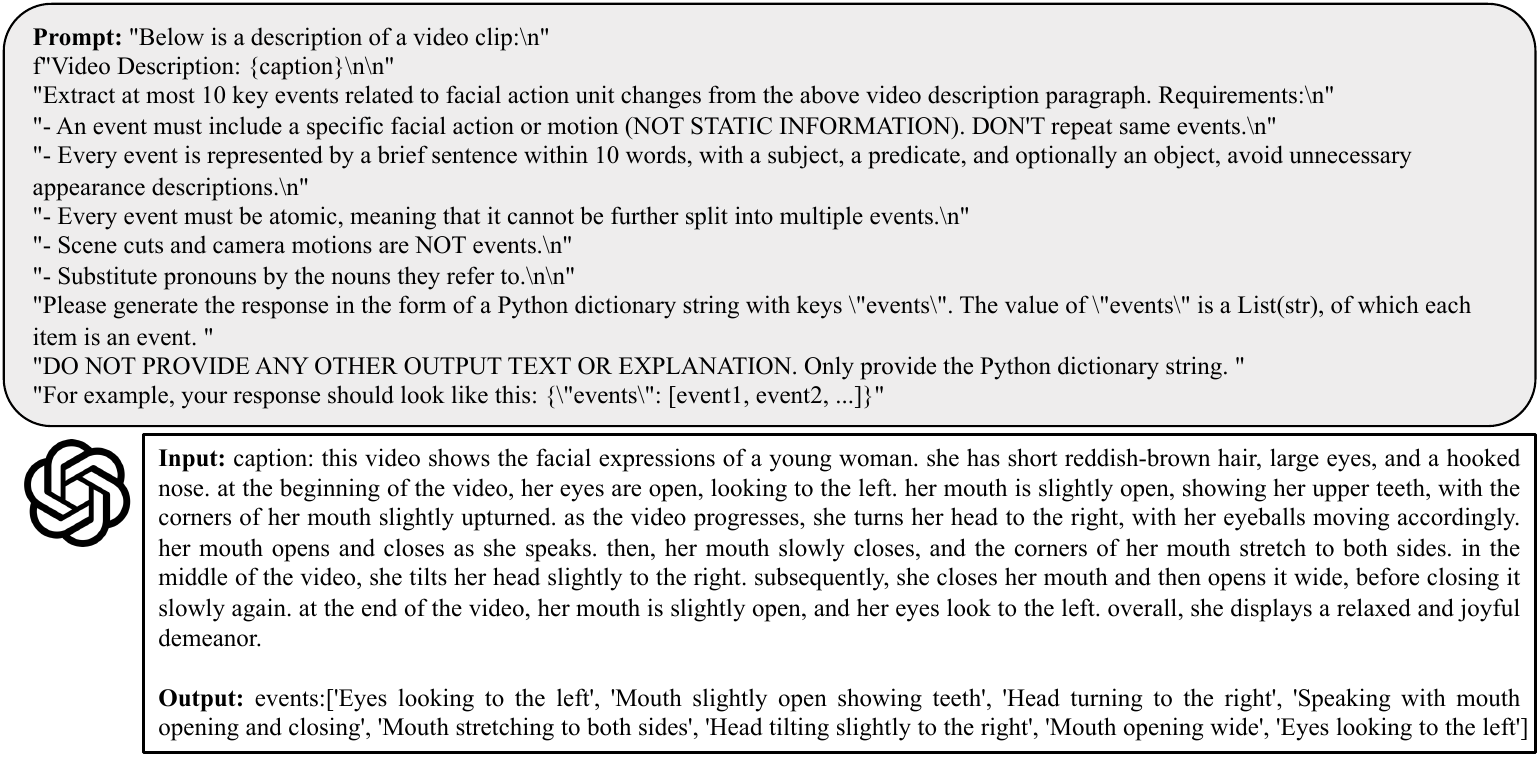}
    \vspace{-0.1cm}
    \caption{\textbf{Event extraction.} We use ChatGPT with the specific prompt to extract event from video captions. \vspace{-0.3cm}}
    \label{fig: extract} 
\end{figure*}

In Figure~\ref{fig: extract}, we present the entire process of event extraction. We used prompts similar to those in AutoDQ~\cite{wang2024tarsierrecipestrainingevaluating} for event extraction, but with a focus on events related to facial movements and expressions rather than general events. Specifically, our prompts are designed to guide ChatGPT in identifying and extracting relevant events from the input text. The input to the system consists of either generated text or reference text, and the output from ChatGPT is a structured list of events. Each event in the list includes a description of the facial movement or expression. This focused approach ensures that the extracted events are highly relevant to the specific task of evaluating facial expressions and movements.

\begin{figure*}

    \centering
    \includegraphics[width=0.98\linewidth]{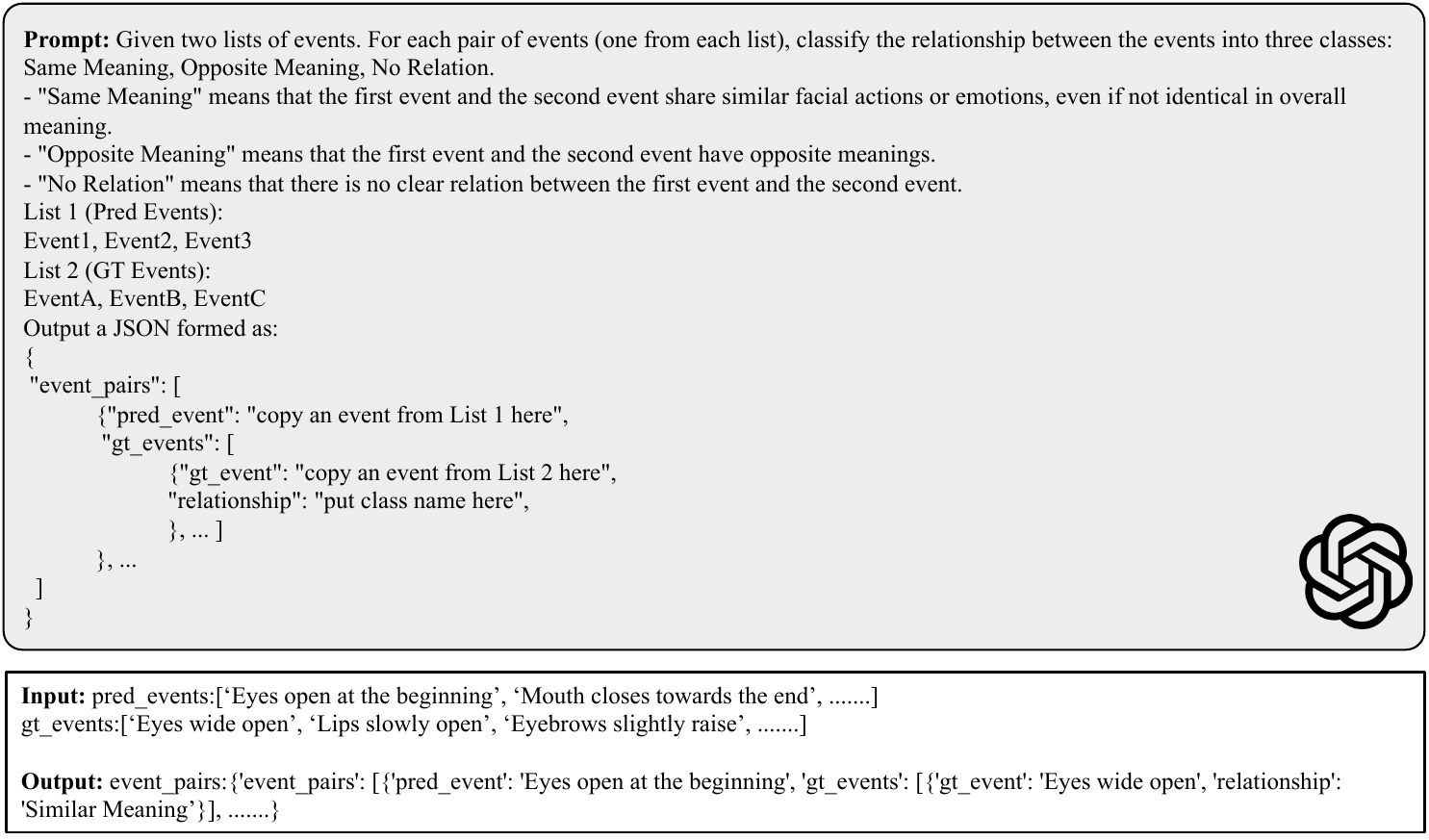}
    \vspace{-0.1cm}
    \caption{\textbf{Event Matching.} After event extraction, We further use ChatGPT with the specific prompt to match event from predicted events and ground-truth events.\vspace{-0.3cm}}
    \label{fig: event match} 
\end{figure*}

In Fig.~\ref{fig: event match}, we present the process of event relation classification and the prompts used. In this section, we rely on carefully designed prompts to use ChatGPT for classifying the relationships between generated events and reference events into three categories: Same Meaning, Opposite Meaning, and No Relation.

\subsection{VideoChatGPT Scores}
\begin{figure*}

    \centering
    \includegraphics[width=0.95\linewidth]{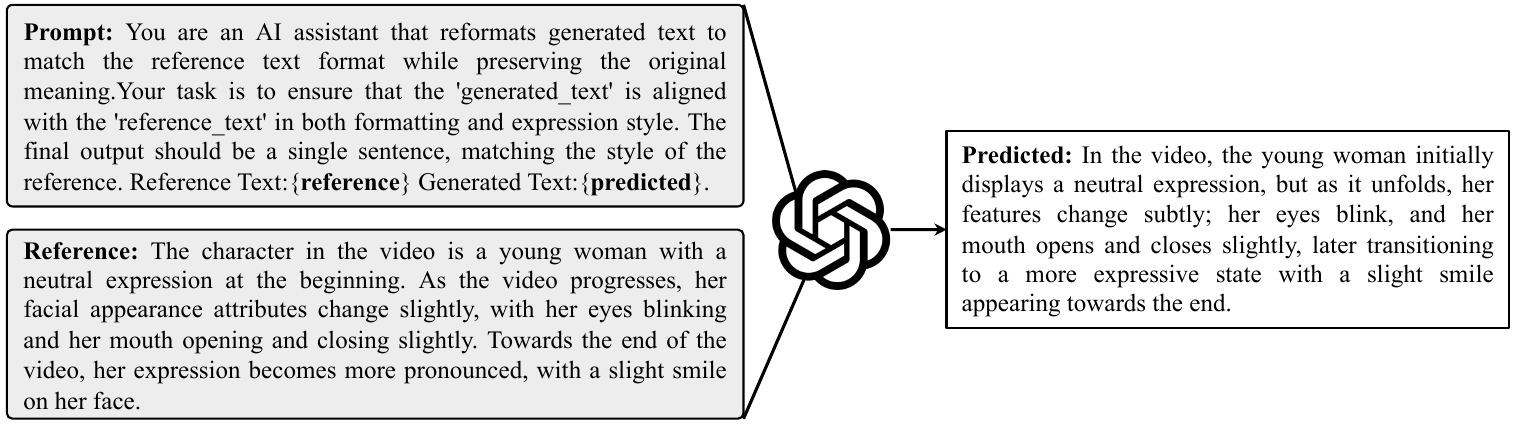}
    \vspace{-0.1cm}
    \caption{\textbf{Format alignment before calculating the VideoChatGPT scores.}\vspace{-0.3cm}}
    \label{fig: align} 
\end{figure*}

To ensure a fair comparison of the scores reported by VideoChatGPT~\cite{maaz2024videochatgptdetailedvideounderstanding}, we used ChatGPT to align the format of the results generated by different methods. Specifically, we standardized the output formats to ensure consistency in how the results are presented. This alignment process involved converting the outputs from various methods into a unified format, which allowed for a more direct and accurate comparison. The detailed process, along with the results before and after alignment, is shown in Fig.~\ref{fig: align}.
\section{Instructions in our FDA}
\begin{figure*}

    \centering
    \includegraphics[width=0.95\linewidth]{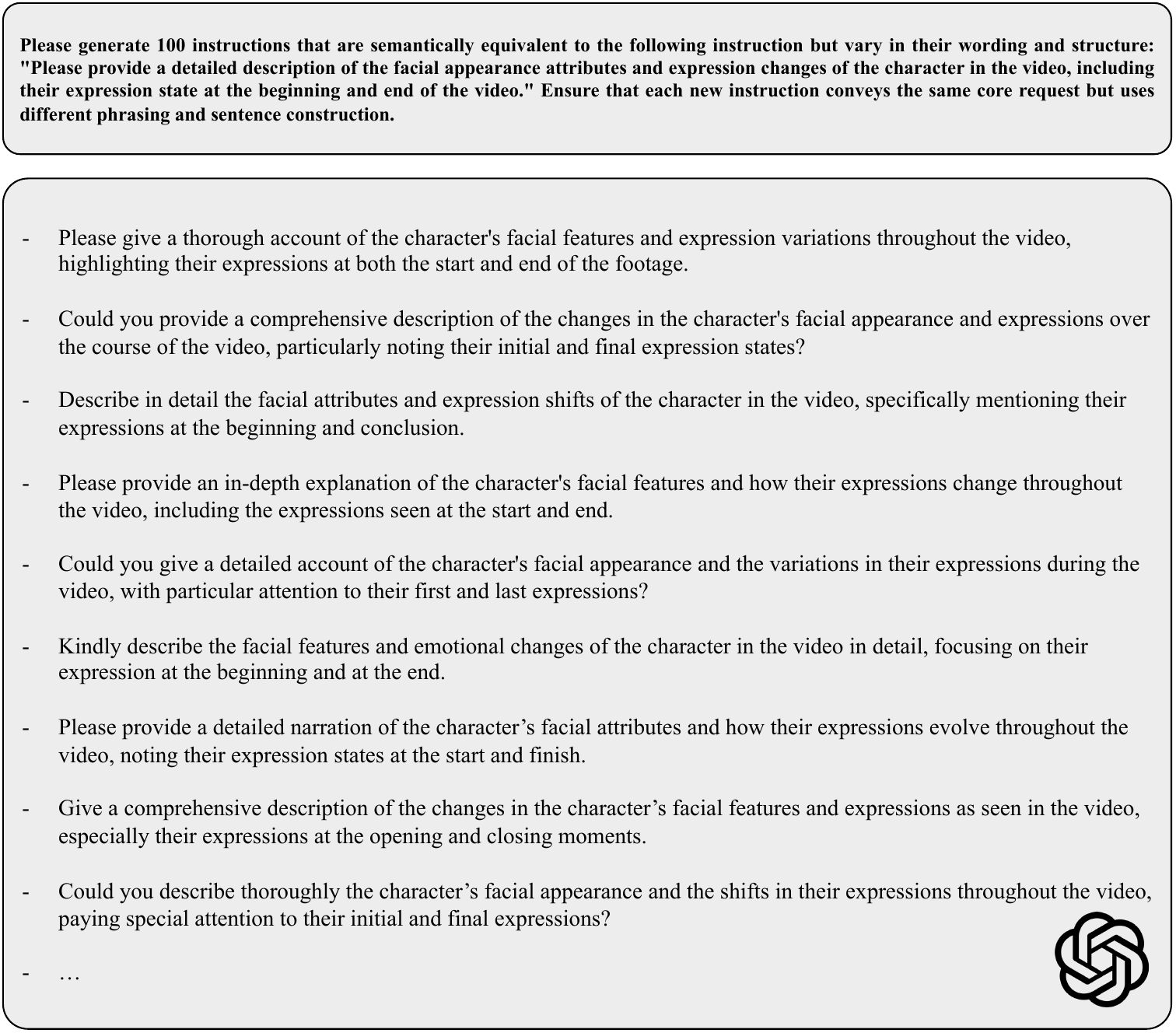}
    \vspace{-0.1cm}
    \caption{\textbf{Prompt used to generate instructions and some instances of FDA instructions.}\vspace{-0.3cm}}
    \label{fig:instruction} 
\end{figure*}

To ensure the diversity of instructions, we used ChatGPT to generate 100 different phrasings of the same instruction. From these, we manually selected 20 instructions to ensure a wide range of variations and maintain the quality of the final set. In Fig.~\ref{fig:instruction}, we illustrate this process, including the prompts we used and some of the generated instructions.

\section{More visualizations of FDA}
\begin{figure*}
\vspace{-5mm}
    \centering
    \includegraphics[width=0.98\linewidth]{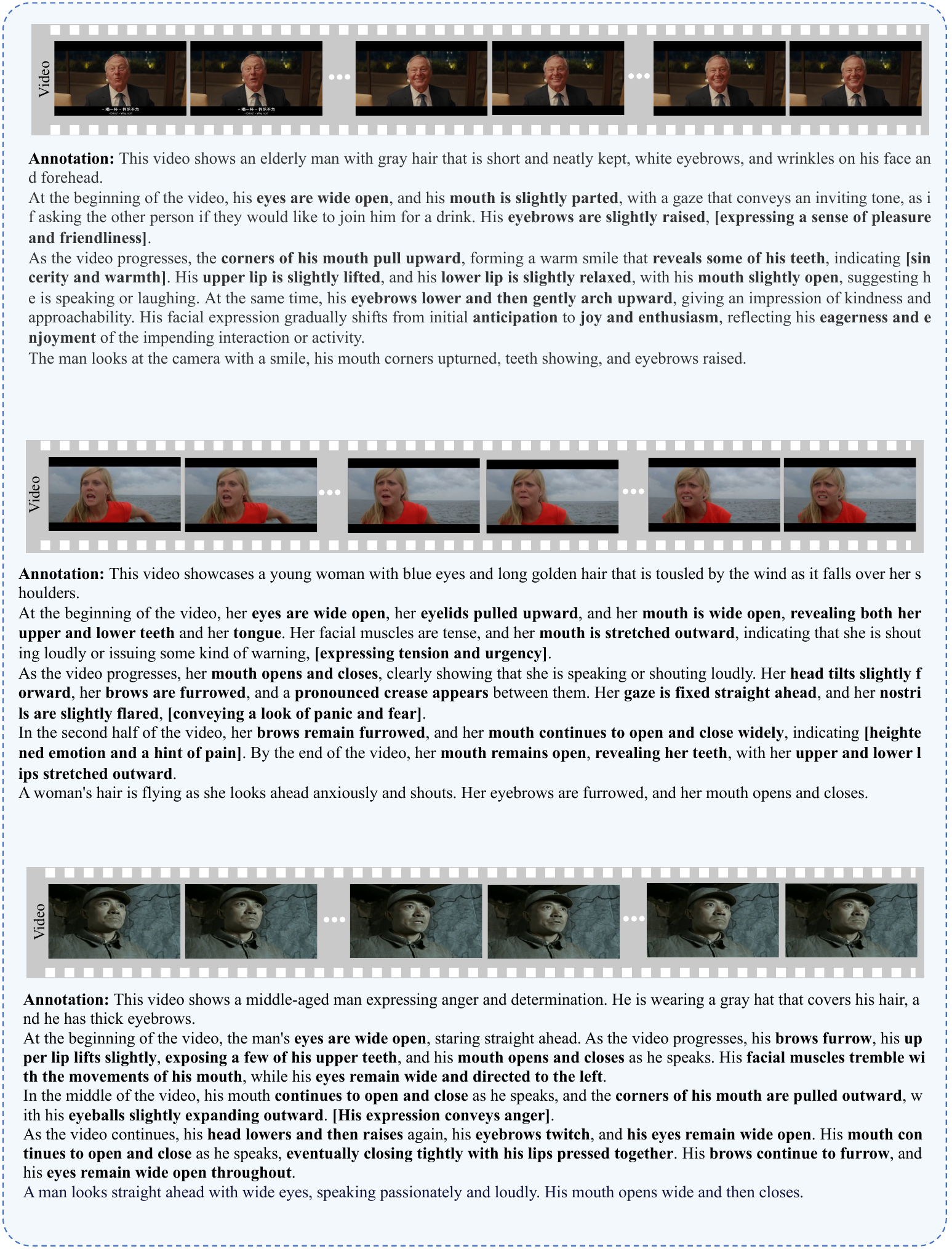}
    \vspace{-0.1cm}
    \caption{\textbf{More visualizations of dataset annotations.}\vspace{-0.3cm}}
    \label{fig:sample_1} 
\end{figure*}

\begin{figure*}
\vspace{-5mm}
    \centering
    \includegraphics[width=0.98\linewidth]{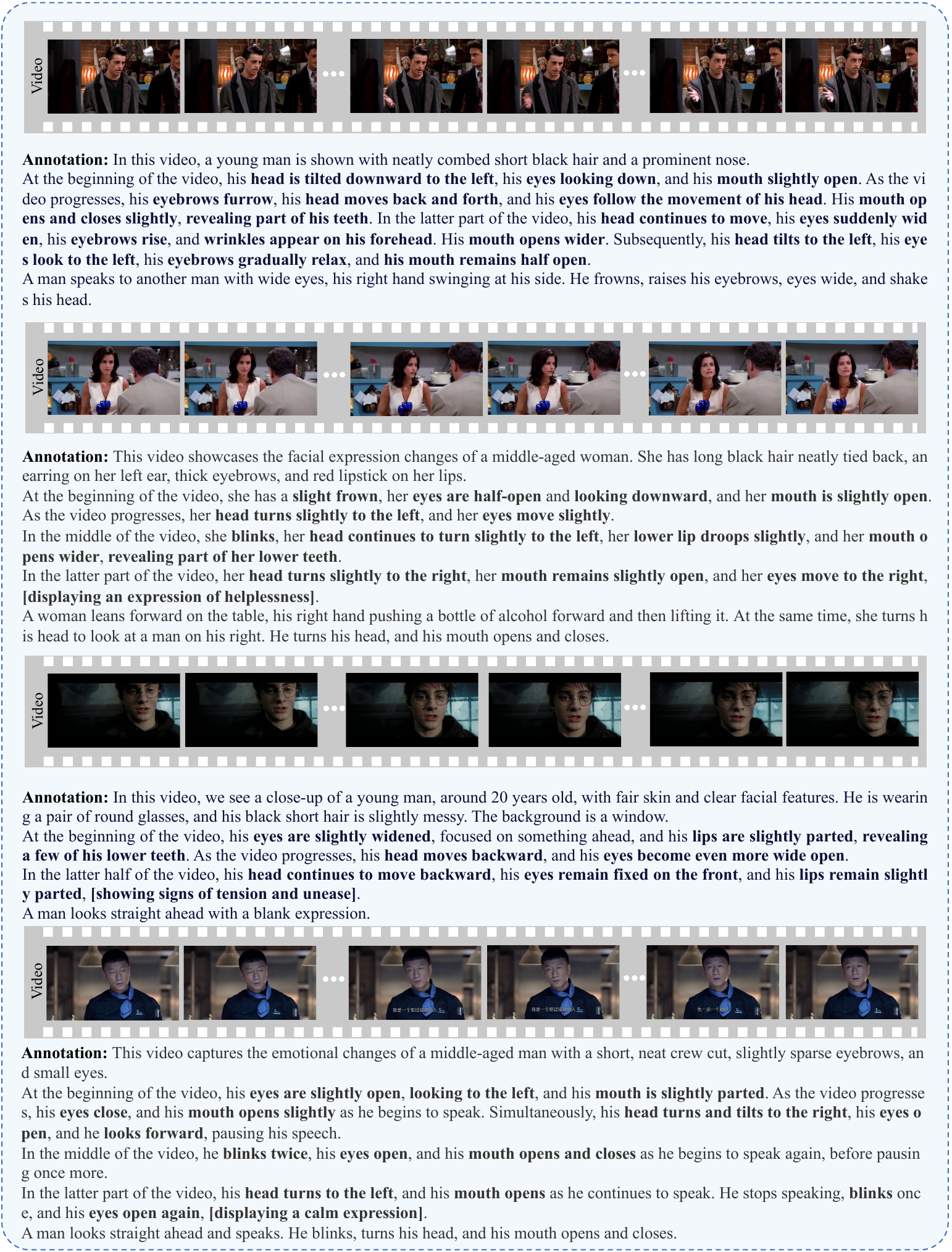}
    \vspace{-0.1cm}
    \caption{\textbf{More visualizations of dataset annotations.}\vspace{-0.3cm}}
    \label{fig:sample_2} 
\end{figure*}

\begin{figure*}
    \vspace{-5mm}
    \centering
    \includegraphics[width=0.95\linewidth]{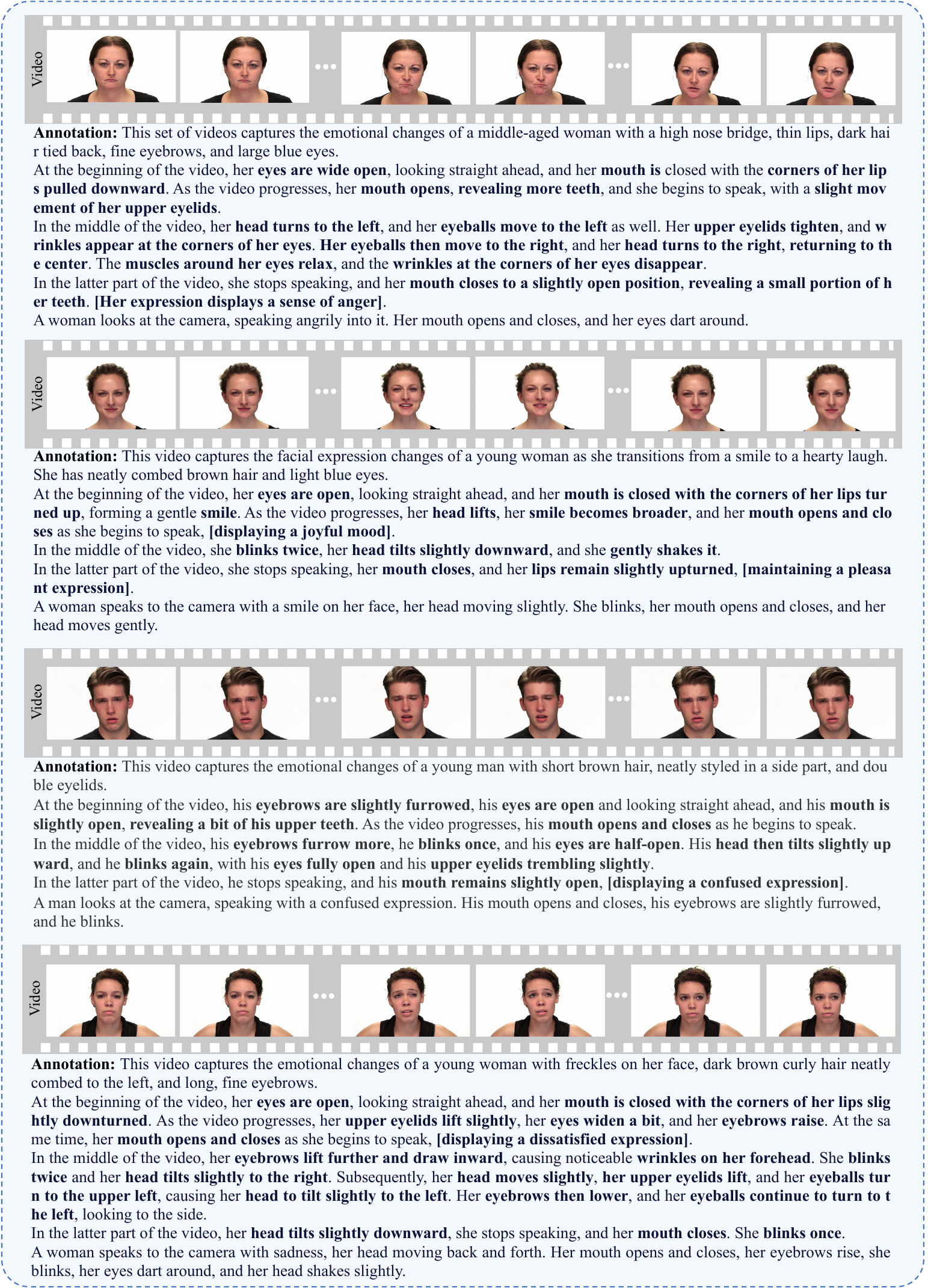}
    \vspace{-0.1cm}
    \caption{\textbf{More visualizations of dataset annotations.}\vspace{-0.3cm}}
    \label{fig:sample_3} 
\end{figure*}

In Fig.~\ref{fig:sample_1}, Fig.~\ref{fig:sample_2}, and Fig.~\ref{fig:sample_3}, we present additional sample data from the datasets. Consistent with the description in Sec.~\ref{sec:dataset}, our annotation content primarily consists of three parts: descriptions of character appearance, descriptions of facial changes, and overall content descriptions. The core of our work lies in the facial change descriptions, where we provide detailed accounts of the facial changes of the main characters in the video. We retain reasonable inferences generated during the pre-generation phase, marked with special symbols to distinguish them from directly observed data.

\clearpage

\begin{table*}[htbp]

\centering
\renewcommand{\arraystretch}{1.6} 
\setlength{\tabcolsep}{5pt} 
\begin{tabularx}{\textwidth}{|l|c|X|X|c|c|c|}
\hline
\textbf{Database} & \textbf{\#Sample} & \textbf{Source} & \textbf{Expression annotation} & \textbf{In-the-wild?} & \textbf{Annotation Times} & \textbf{Modality} \\ \hline
CK+~\cite{CK+}  & 327 & Lab & 6 expressions + neutral and contempt & No & - & Video \\ \hline
MMI~\cite{MMI} & 2900 & Lab & 6 expressions + neutral & No & - & Video \\ \hline
BP4D~\cite{ZHANG2014692} & 328 & Lab & 6 expressions + embarrassment and pain & No & - & Video+Audio \\ \hline
Aff-Wild2~\cite{kollias2019expressionaffectactionunit} & 84 & Web + YouTube & 6 expressions + neutral & Yes & 3 & Video+Audio \\ \hline
AFEW 7.0~\cite{KOSSAIFI201723} & 1,809 & 54 movies & 6 expressions + neutral & Yes & 2 & Video+Audio \\ \hline
CAER~\cite{lee2019context} & 13,201 & 79 TV dramas & 6 expressions + neutral & Yes & 3 & Video+Audio \\ \hline
EmoVoxCeleb~\cite{albanie2018emotionrecognitionspeechusing} & 22,496 & Interview videos from YouTube & 6 expressions+neutral and contempt & Yes & Auto & Video+Audio \\ \hline
DFEW~\cite{jiang2020dfewlargescaledatabaserecognizing} & 16,372 & 1500 movies & 6 expressions + neutral & Yes & 10 & Video+Audio \\ \hline
MAFW~\cite{liu2023mafwlargescalemultimodalcompound} & 10,045 & 1,600 movies + TV dramas, 20,000 short videos from reality shows, talk shows, news, etc, 2,045 clips & 11 single expressions + 32 multiple expressions + emotional descriptive text & Yes & 11 & Video+Audio+Text \\ \hline
Our FDA & 5033 & MFAW~\cite{liu2023mafwlargescalemultimodalcompound} + DFEW~\cite{jiang2020dfewlargescaledatabaserecognizing} + AFEW~\cite{KOSSAIFI201723} + MER~\cite{GomezCanon2021SPM} + CAER~\cite{lee2019context} + FERV39K \cite{wang2022ferv39klargescalemultiscenedataset} + RAVEDESS \cite{Ravedess} + Pexels video + movie clips & 10.3 expression-related tokens per clip & Yes & 3 & Video+Audio+Text \\ \hline
\end{tabularx}
\caption{Comparison with existing video expression datasets.}
\label{tabel-comp}
\end{table*}

\newpage

\newpage

\clearpage
{
    \small
    \bibliographystyle{ieeenat_fullname}
    \bibliography{main}
}


\end{document}